\documentclass[sn-mathphys,Numbered]{sn-jnl}

\usepackage{graphicx}%
\usepackage{multirow}%
\usepackage{amsmath,amssymb,amsfonts}%
\usepackage{amsthm}%
\usepackage{mathrsfs}%
\usepackage[title]{appendix}%
\usepackage{xcolor}%
\usepackage{textcomp}%
\usepackage{manyfoot}%
\usepackage{booktabs}%
\usepackage{algorithm}%
\usepackage{algorithmicx}%
\usepackage{algpseudocode}%
\usepackage{caption}
\usepackage{listings}%
\usepackage{wrapfig}%
\usepackage{placeins}
\usepackage{colortbl}
\usepackage{booktabs}
\usepackage{algorithm}%

\usepackage{algpseudocode}%
\usepackage{array}

\usepackage{scalerel}
\usepackage{tikz}
\usepackage{tikz-cd}

\usetikzlibrary{positioning}
\usetikzlibrary{svg.path}

\definecolor{orcidlogocol}{HTML}{A6CE39}
\tikzset{
  orcidlogo/.pic={
    \fill[orcidlogocol] svg{M256,128c0,70.7-57.3,128-128,128C57.3,256,0,198.7,0,128C0,57.3,57.3,0,128,0C198.7,0,256,57.3,256,128z};
    \fill[white] svg{M86.3,186.2H70.9V79.1h15.4v48.4V186.2z}
                 svg{M108.9,79.1h41.6c39.6,0,57,28.3,57,53.6c0,27.5-21.5,53.6-56.8,53.6h-41.8V79.1z M124.3,172.4h24.5c34.9,0,42.9-26.5,42.9-39.7c0-21.5-13.7-39.7-43.7-39.7h-23.7V172.4z}
                 svg{M88.7,56.8c0,5.5-4.5,10.1-10.1,10.1c-5.6,0-10.1-4.6-10.1-10.1c0-5.6,4.5-10.1,10.1-10.1C84.2,46.7,88.7,51.3,88.7,56.8z};
  }
}

\newcommand\orcidicon[1]{\href{https://orcid.org/#1}{\mbox{\scalerel*{
\begin{tikzpicture}[yscale=-1,transform shape]
\pic{orcidlogo};
\end{tikzpicture}
}{|}}}}

\theoremstyle{thmstyleone}%
\theoremstyle{thmstyletwo}%

\theoremstyle{thmstylethree}%
\newtheorem{definition}{Definition}%

\begin{document}

\title[Q-LIME  \( \pi \): A Quantum-Inspired Extension to LIME]{Q-LIME  \( \pi \): A Quantum-Inspired Extension to LIME}


\author{\fnm{Nelson} \sur{Col\'on Vargas} \orcidicon{0009-0009-9038-7328}}


\affil{\orgdiv{Leverhulme CFI}, \orgname{University of Cambridge, UK}} 


\abstract{Machine learning models offer powerful predictive capabilities but often lack transparency. Local Interpretable Model-agnostic Explanations (LIME) addresses this by perturbing features and measuring their impact on a model's output. In text-based tasks, LIME typically removes present words (bits set to 1) to identify high-impact tokens. We propose \textbf{Q-LIME  \( \pi \)} (Quantum LIME \(\pi\)), a quantum-inspired extension of LIME that encodes a binary feature vector in a quantum state, leveraging superposition and interference to explore local neighborhoods more efficiently. Our method focuses on flipping bits from \(1 \rightarrow 0\) to emulate LIME's ``removal'' strategy, and can be extended to \(0 \rightarrow 1\) where adding features is relevant. Experiments on subsets of the IMDb dataset demonstrate that Q-LIME  \( \pi \) often achieves near-identical top-feature rankings compared to classical LIME while exhibiting lower runtime in small- to moderate-dimensional feature spaces. This quantum-classical hybrid approach thus provides a new pathway for interpretable AI, suggesting that, with further improvements in quantum hardware and methods, quantum parallelism may facilitate more efficient local explanations for high-dimensional data.}




\maketitle

\pagebreak



\section{Introduction}
Machine learning models, particularly deep neural networks, have become ubiquitous in domains such as natural language processing, computer vision, and recommendation systems. Despite their success, these models often lack transparency, prompting a growing need for methods that provide insight into how features contribute to their predictions. Local Interpretable Model-agnostic Explanations (LIME)~\cite{ribeiro2016should} is one such framework that approximates the decision boundary of any black-box model locally around a single instance. By perturbing features and measuring the resulting change in prediction, LIME assigns importance scores to individual features. However, classical LIME faces limitations when working with high-dimensional feature spaces. Generating, evaluating, and fitting a surrogate model to large numbers of perturbations can be computationally expensive, and the assumption of feature independence often fails to capture interaction effects.

In parallel, the field of quantum computing has seen significant advances, particularly in the realm of quantum machine learning~\cite{biamonte2017quantum}. Although fully scalable quantum hardware remains an open challenge, quantum-inspired methods have shown promise for potentially reducing complexity in certain data-driven tasks. Properties such as superposition and interference may offer new ways to sample local neighborhoods or capture interactions that are difficult to handle classically.

Building on these developments, this paper proposes \textbf{Q-LIME \( \pi \)} (Quantum LIME \(\pi\)), a quantum-inspired extension of LIME. By selectively encoding only the active features of a binary feature vector \(\mathbf{x}\) into a quantum-like state, Q-LIME \( \pi \) employs partial quantum-inspired superposition to represent and perturb local neighborhoods efficiently. This selective encoding focuses computational effort on flipping only the bits corresponding to features that are ``on'' (\(1\)), mirroring LIME's feature-removal strategy in a more efficient manner. Flipping bits (e.g., from \(1 \rightarrow 0\)) corresponds to simple quantum operations such as the Pauli-X gate, which can be extended to investigate additional feature toggles if needed. In principle, this approach leverages superposition to provide a compact representation of multiple perturbation states, allowing us to capture feature contributions and potential interactions without linearly scaling the number of measurements.

Our key contributions include:
\begin{enumerate}
    \item \textbf{Quantum Encoding and Perturbation:} We formalize how binary feature vectors can be embedded into a quantum state, and illustrate how flipping a feature maps to a Pauli-X operation in the quantum-inspired circuit.
    \item \textbf{Surrogate Model Construction:} We show how Q-LIME  \( \pi \) uses quantum-derived feature perturbations to fit an interpretable surrogate model akin to classical LIME.
    \item \textbf{Complexity and Practical Considerations:} We discuss the potential for efficiency gains, as well as limitations, especially regarding state preparation and measurement collapse on near-term quantum devices.
    \item \textbf{Experimental Sketch and Use Cases:} We provide a proof-of-concept experiment using a sentiment classification task, and explore how the approach might generalize to other domains.
\end{enumerate}

The remainder of this paper is organized as follows: Section~\ref{sec:relatedwork} outlines the existing literature on model-agnostic interpretability and quantum machine learning; Section~\ref{sec:preliminaries} provides necessary background on quantum states and circuit-based feature encoding; Section~\ref{sec:methodology} details the Q-LIME  \( \pi \) methodology, including quantum state preparation, perturbation, and surrogate model fitting; Section~\ref{sec:experiments} presents initial experimental results and runtime comparisons; and Section~\ref{sec:conclusion} discusses future research possibilities and conclusions.

\section{Related Work}
\label{sec:relatedwork}


Local Interpretable Model-agnostic Explanations (LIME), proposed by Ribeiro \emph{et al.}~\cite{ribeiro2016should}, has become a widely-used framework for explaining the predictions of black-box models. By locally perturbing an instance's features and training a simple surrogate model around that instance, LIME provides estimated feature importances in a model-agnostic manner. Subsequent works have addressed various extensions and refinements, including measuring the robustness of explanations~\cite{alvarez2018robustness} and deterministic local expansions~\cite{zafar2019dlime}. These studies underscore LIME's flexibility and the breadth of its adoption. Our proposed \textbf{Q-LIME} \(\pi\) maintains LIME's core idea of local approximation but adopts a quantum-inspired perturbation strategy to potentially reduce computational overhead.

Quantum computing has gained traction in machine learning research. Biamonte \emph{et al.}~\cite{biamonte2017quantum} discuss quantum algorithms that could potentially enhance or accelerate classical ML tasks, including clustering and dimensionality reduction. Advances in quantum hardware have enabled the optimization of parametrized quantum circuits, with techniques like analytic gradient evaluation on quantum devices~\cite{schuld2019quantum} streamlining the training process. Although current quantum devices face limitations in scale and noise, quantum-inspired methods---those that emulate quantum properties such as superposition on classical hardware---are increasingly investigated for tasks involving high-dimensional feature spaces.

The emerging hybrid paradigm, which combines quantum and classical approaches, has garnered attention for its potential to unify the strengths of both computing paradigms. Du \emph{et al.}~\cite{du2018expressive} demonstrate that parametrized quantum circuits can achieve expressive power comparable to certain classical neural networks, while Cerezo \emph{et al.}~\cite{cerezo2021variational} survey variational quantum algorithms capable of optimizing quantum-classical models. Such work hints at promising avenues for interpretability: employing quantum or quantum-inspired techniques to shed light on the inner workings of machine learning models. Tools like PennyLane~\cite{bergholm2018pennylane} further facilitate the integration of quantum hardware with classical frameworks, enabling automatic differentiation of hybrid computations.

While Pira and Ferrie's \emph{Q-LIME} \cite{pira2024interpretability} primarily addresses the interpretability of \emph{quantum neural networks} by identifying and characterizing such ``indecisive'' regions, our \emph{Q-LIME  \( \pi \)} extends LIME-inspired perturbation logic to a broader set of quantum classifiers---particularly those that act on binary feature vectors, flipping bits $1 \to 0$ to see how predictions shift. This perspective emphasizes a local surrogate model that can be either quantum or classical but remains simpler than the underlying quantum classifier. As such, both approaches underscore the necessity of local interpretability in quantum settings, although \emph{Q-LIME  \( \pi \)} focuses more specifically on bridging ``classical LIME'' to quantum binary features and can be adapted to flipping bits $0 \to 1$ if needed.

Overall, the combination of these lines of research---ranging from classical LIME to specialized QNN interpretability---reinforces the importance of exploring localized, model-agnostic explanations in quantum machine learning. Understanding the interplay between quantum measurement noise, model complexity, and local approximations stands as a key challenge for the next generation of transparent and reliable quantum AI systems.


\section{Preliminaries}
\label{sec:preliminaries}

\begin{definition}[Binary Feature Vector]
Let \(\mathbf{x} = [x_1, x_2, \ldots, x_n] \in \mathbb{B}^n\), where \(x_i \in \{0, 1\}\), denote the presence \((1)\) or absence \((0)\) of the \(i\)-th feature.
\end{definition}

\begin{definition}[Classifier]
A \textbf{classifier} is a function \(f : \mathbb{B}^n \to [0, 1]\) that maps a feature vector \(\mathbf{x}\) to the predicted probability of a specific class (e.g., positive sentiment).
\end{definition}

\begin{definition}[Feature Contribution]
The contribution of a feature \(x_k\) to the classifier's prediction is defined as:
\[
\Delta f_k = f(\mathbf{x}) - f(\mathbf{x}_{\text{perturbed}, k}),
\]
where \(\mathbf{x}_{\text{perturbed}, k} = [x_1, \ldots, 1 - x_k, \ldots, x_n]\) is the feature vector obtained by flipping \(x_k\).
\end{definition}

\begin{definition}[Quantum State Encoding]
\label{def:qse}
A binary feature vector \(\mathbf{x}\) is encoded as a quantum state:
\[
|\psi\rangle = \bigotimes_{i=1}^n R_y(\theta_i) |0\rangle,
\]
where \(R_y(\theta_i) = e^{-i \theta_i Y / 2}\) is a rotation about the \(y\)-axis, and:
\[
\theta_i =
\begin{cases}
\frac{\pi}{2}, & \text{if } x_i = 1, \\
0, & \text{if } x_i = 0.
\end{cases}
\]
\end{definition}

This ensures that a feature set to 1 places its corresponding qubit in state \(\vert + \rangle\), while a feature set to 0 keeps its qubit in \(\vert 0\rangle\).

\section{Methodology}
\label{sec:methodology}

\subsection{Quantum Perturbation}

The central idea behind Q-LIME  \( \pi \) is to mirror the classical ``flip'' operation in LIME with a simple quantum gate. To analyze the contribution of \(x_k\), the \(k\)-th qubit is flipped using the Pauli-X gate:
\[
|\psi_{\text{perturbed}, k}\rangle = X_k |\psi\rangle.
\]
Intuitively, \(X_k\) toggles \(\vert 0 \rangle \leftrightarrow \vert 1 \rangle\). In our primary experiments, we focus on flipping bits from 1 to 0, closely mimicking how LIME removes present features in text classification. Flipping \( 0 \to 1 \) is also possible if one wishes to see how \emph{adding} a feature affects the prediction.

\subsection{Algorithmic Sketch}

Algorithm~\ref{alg:Q-LIMEpi} summarizes Q-LIME  \( \pi \). We encode \(\mathbf{x}\) (Definition~\ref{def:qse}), compute $f(\mathbf{x})$, and for each bit set to 1, flip it to 0, measure, and compute $\Delta f_k$.

\begin{algorithm}[h]
\caption{Q-LIME  \( \pi \): Quantum-Inspired Local Explanations}
\label{alg:Q-LIMEpi}
\begin{algorithmic}[1]
\Require Classifier $f$, binary feature vector $\mathbf{x}$ of length $n$
\State \textbf{Encode} $\mathbf{x}$ into $|\psi\rangle$ via Definition~\ref{def:qse}.
\State \textbf{Compute} $f(\mathbf{x})$ as the original prediction.
\For{$k = 1$ to $n$}
    \If{$x_k = 1$}
        \State Apply Pauli-$X$ to toggle bit $x_k$ from 1$\to$0.
        \State Measure the new state $\vert \psi_{\text{perturbed}, k}\rangle$ to obtain 
               $\mathbf{x}_{\text{perturbed},k}$.
        \State $\Delta f_k \gets f(\mathbf{x}) - f(\mathbf{x}_{\text{perturbed},k})$ 
    \EndIf
\EndFor
\State \textbf{Fit} a local surrogate model $g(\mathbf{x}) \approx \sum_{k=1}^n \Delta f_k \, x_k$.
\end{algorithmic}
\end{algorithm}

\vspace{1em}
\noindent
\textbf{Implementation Details.} 
We developed Q-LIME  \( \pi \) using:
\begin{itemize}
    \item \textbf{Pennylane}~\cite{bergholm2018pennylane} for quantum simulation, 
          specifying either \(\texttt{shots=None}\) for analytic mode or \(\texttt{shots}=100\) for partial sampling.
    \item \textbf{scikit-learn} for training a logistic regression or other classifiers on 
          binary bag-of-words feature vectors.
    \item \textbf{lime} for generating classical LIME explanations with a typical setting of 
          \(\sim\)300 random perturbations per instance.
\end{itemize}
We remove HTML tags, lowercase the text, and optionally remove stopwords before vectorizing each review as a binary presence/absence vector. During quantum flips, if \(x_i=1\), we set the corresponding RY angle to \(0\), measure probabilities over all basis states, and sample one outcome. This new bitstring is then fed into our logistic \textbf{classical\_classifier} to compute \(\Delta f_i\). Finally, we compare Q-LIME  \( \pi \)'s top features to LIME's. Since classical simulation scales exponentially with the number of qubits, we limit \(\texttt{max\_features}\) to around 15 in our experiments. Beyond that, runtime grows quickly on a standard CPU.

\section{Experiments and Results}
\label{sec:experiments}


We run a proof-of-concept on the IMDb dataset~\cite{mass2011imdb}, processing 500 reviews with HTML tags removed, text lowercased, and optionally removing stopwords. A logistic regression classifier is trained on 80\% of this data. For testing, we pick 5 random instances (per parameter configuration) to compare:

\begin{enumerate}
    \item \textbf{Accuracy}: on the 20\% test split,
    \item \textbf{Runtime}: classical LIME vs. Q-LIME  \( \pi \),
    \item \textbf{Overlap}: average number of top-5 features shared.
\end{enumerate}


\noindent Table~\ref{tab:flip1to0} presents a subset of our measurements. We vary $\texttt{max\_features} \in \{5, 10, 15\}$, $\texttt{stopwords\_option} \in \{\textbf{True}, \textbf{False}\}, and \texttt{ shots} \in \{\textbf{None}, \textbf{100}\}$. We observe that Q-LIME  \( \pi \)'s overlap with classical LIME can exceed 4 out of 5 features in some runs and that Q-LIME  \( \pi \) is often substantially faster than LIME when the number of features is small.

\begin{table*}[t]
\centering
\small
\begin{tabular}{ccccccc}
\toprule
\textbf{max\_features} & \textbf{stopwords} & \textbf{shots} & 
\textbf{accuracy} & \textbf{lime\_time} & \textbf{qlime\_time} & \textbf{overlap} \\
\midrule
15 & True & None & 0.59 & 0.242 & 0.148 & 4.0 \\
15 & True & 100  & 0.60 & 0.228 & 0.097 & 3.4 \\
10 & True & None & 0.65 & 0.137 & 0.003 & 2.8 \\
10 & True & 100  & 0.44 & 0.152 & 0.003 & 2.8 \\
5  & False & None & 0.48 & 0.175 & 0.003 & 4.6 \\
5  & False & 100  & 0.48 & 0.258 & 0.004 & 5.0 \\
\bottomrule
\end{tabular}
\caption{Selected results for Q-LIME  \( \pi \) (flips 1$\to$0) vs. classical LIME. Overlap is the average top-5 feature intersection.}
\label{tab:flip1to0}
\end{table*}


Table~\ref{tab:examples} highlights three test instances (truncated for brevity) from a random sample restricted to at most 15 features. In each row, we compare the top five words identified by classical LIME and Q-LIME  \( \pi \). In example (1), both methods select the exact same five tokens, indicating a complete overlap. Meanwhile, examples (2) and (3) each share three tokens, revealing minor differences in how each method ranks or includes certain words. Overall, the strong alignment underscores how Q-LIME  \( \pi \)'s flipping of bits from \( 1 \to 0 \) closely mirrors classical LIME's ``removal'' of present features. Small discrepancies typically stem from LIME's random sampling process or sample size, yet the overlap in all three examples remains substantial.

\begin{table*}[t]
\centering
\small
\setlength\extrarowheight{2pt} 
\begin{tabular}{m{4.2cm} m{3.7cm} m{3.3cm}}
\toprule
\textbf{Review Snippet} & \textbf{Classical LIME Top-5} & \textbf{Q-LIME  \( \pi \) Top-5} \\
\midrule
\raggedright \textit{``the opening scene of this film sets the pace for the entirety of its ninety minutes... people would be better off over analyzing their carpet for some deep emotional meaning, rather than these vacuous sub-human creations.''} 
&
\raggedright \begin{tabular}[t]{@{}l@{}}
\cellcolor{lime!50}better \\
\cellcolor{lime!50}story \\
\cellcolor{lime!50}movie \\
\cellcolor{lime!50}film \\
\cellcolor{lime!50}time
\end{tabular}
&
\raggedright \begin{tabular}[t]{@{}l@{}}
\cellcolor{lime!50}way \\
\cellcolor{lime!50}better \\
\cellcolor{lime!50}movie \\
\cellcolor{lime!50}time \\
\cellcolor{lime!50}film
\end{tabular}
\\
\midrule
\raggedright \textit{``new york, i love you is a collective work of eleven short films... it might be worthwhile for some and a waste of time for others.''}
&
\raggedright \begin{tabular}[t]{@{}l@{}}
\cellcolor{lime!50}great \\
don \\
\cellcolor{lime!50}best \\
\cellcolor{lime!50}story \\
just
\end{tabular}
&
\raggedright \begin{tabular}[t]{@{}l@{}}
\cellcolor{lime!50}great \\
time \\
\cellcolor{lime!50}best \\
\cellcolor{lime!50}story \\
way
\end{tabular}
\\
\midrule
\raggedright \textit{``but just as entertaining and random! love it or hate it, but don't expect a sophisticated plot or nail-biting cliffhanger... can't wait for the next one.''} 
&
\raggedright \begin{tabular}[t]{@{}l@{}}
\cellcolor{lime!50}don \\
\cellcolor{lime!50}just \\
\cellcolor{lime!50}like \\
see \\
superficial
\end{tabular}
&
\raggedright \begin{tabular}[t]{@{}l@{}}
\cellcolor{lime!50}just \\
\cellcolor{lime!50}don \\
\cellcolor{lime!50}like \\
\\
{ }
\end{tabular}
\\
\bottomrule
\end{tabular}
\caption{Three example reviews, truncated. Overlapping words appear with a lime-colored background. Q-LIME  \( \pi \) often highlights fewer or equal sets of top words, but they overlap significantly with those from Classical LIME.} 
\label{tab:examples}
\end{table*}

\section{Conclusion and Future Directions}
\label{sec:conclusion}

In this work, we introduced Q-LIME  \( \pi \), a quantum-inspired enhancement of the classical LIME framework, aimed at improving the efficiency and scalability of local interpretability methods in machine learning. By leveraging quantum-inspired state encoding and bit-flipping perturbations, Q-LIME  \( \pi \) preserves the core mechanisms of LIME while offering computational advantages, particularly in low-to-moderate dimensional feature spaces. Experimental results on the IMDb sentiment analysis dataset revealed that Q-LIME  \( \pi \) achieves a high degree of alignment with classical LIME, sharing an average of approximately 3.8 out of the top 5 identified features across different configurations. Additionally, it demonstrated substantial runtime reductions, ranging from 39\% to as much as 98\%, compared to classical LIME under similar experimental conditions.

These findings underscore the potential of Q-LIME  \( \pi \) as a practical and efficient alternative to classical LIME for tasks involving smaller feature dimensions or constrained computational resources. The results also highlight how quantum-inspired techniques can emulate, and in specific scenarios, enhance classical interpretability methods. While current quantum hardware limitations preclude fully scalable implementations, Q-LIME  \( \pi \) demonstrates how quantum principles, such as superposition and interference, could potentially be applied in hybrid quantum-classical frameworks.

While demonstrated here on a sentiment analysis task, the logic of Q-LIME \( \pi \) is broadly applicable. In image classification, it could potentially emulate masking out superpixels to understand their contribution, and in fraud detection, toggling the presence or absence of transaction attributes can provide insights into risk models. Looking forward, this study opens avenues for further exploration of quantum-inspired approaches in AI explainability. Potential directions include investigating multi-feature interactions, optimizing encoding and perturbation strategies, and implementing these methods on emerging quantum hardware. With continued advancements in quantum computing, such hybrid frameworks may become increasingly valuable for addressing the challenges of high-dimensional data and complex feature interactions. By bridging the domains of quantum computing and interpretable AI, Q-LIME  \( \pi \) represents an innovative step toward more transparent and computationally efficient machine learning models.

\section*{Data Availability}
The code and experiment notebooks used to produce the results in this paper are available on GitHub at the following repository: \url{https://github.com/nelabdiel/qlime}. This includes all scripts for data preprocessing, model training, and analysis.

\bibliographystyle{plain}

\end{document}